\def\year{2022}\relax
\documentclass[letterpaper]{article} % DO NOT CHANGE THIS
\usepackage{aaai22}  % DO NOT CHANGE THIS
\usepackage{times}  % DO NOT CHANGE THIS
\usepackage{helvet}  % DO NOT CHANGE THIS
\usepackage{courier}  % DO NOT CHANGE THIS
\usepackage[hyphens]{url}  % DO NOT CHANGE THIS
\usepackage{graphicx} % DO NOT CHANGE THIS
\usepackage{natbib}  % DO NOT CHANGE THIS AND DO NOT ADD ANY OPTIONS TO IT
\usepackage{caption} % DO NOT CHANGE THIS AND DO NOT ADD ANY OPTIONS TO IT
\DeclareCaptionStyle{ruled}{labelfont=normalfont,labelsep=colon,strut=off} % DO NOT CHANGE THIS
\usepackage{algorithm}
\usepackage{algorithmic}

\usepackage{newfloat}
\usepackage{listings}
\floatstyle{ruled}
\newfloat{listing}{tb}{lst}{}
\floatname{listing}{Listing}
\nocopyright
\usepackage{caption}
\usepackage{subcaption}
\usepackage{siunitx}
\usepackage{booktabs}
\usepackage{amsmath}
\usepackage{amssymb}
\usepackage{multirow}
\usepackage{circledsteps}
\usepackage[capitalise]{cleveref}

\newcommand{\gmean}{our method}
\newcommand{\gmn}{Ours}

\title{Unsupervised Cross-Lingual Transfer of Structured Predictors\\
  without Source Data}
\author {
    Kemal Kurniawan,\textsuperscript{\rm 1}
    Lea Frermann,\textsuperscript{\rm 1}
    Philip Schulz,\textsuperscript{\rm 2}\thanks{Work done outside Amazon.}
    Trevor Cohn\textsuperscript{\rm 1}
}
\affiliations {
    \textsuperscript{\rm 1}School of Computing and Information Systems,
    University of Melbourne\\
    \textsuperscript{\rm 2}Amazon Research\\
    kemal.kurniawan@student.unimelb.edu.au,
    \{lea.frermann,tcohn\}@unimelb.edu.au,
    phschulz@amazon.com
}

\begin{document}

\maketitle

\begin{abstract}
  Providing technologies to communities or domains where training data is scarce
  or protected e.g., for privacy reasons, is becoming increasingly important. To
  that end, we generalise methods for unsupervised transfer from multiple input
  models for structured prediction. We show that the means of aggregating over
  the input models is critical, and that multiplying marginal probabilities of
  substructures to obtain high-probability structures for distant supervision is
  substantially better than taking the union of such structures over the input
  models, as done in prior work. Testing on 18 languages, we demonstrate that
  the method works in a cross-lingual setting, considering both dependency
  parsing and part-of-speech structured prediction problems. Our analyses show
  that the proposed method produces less noisy labels for the distant
  supervision.\footnote{Code: \url{https://github.com/kmkurn/uxtspwsd}}
\end{abstract}

\section{Introduction}

Recent successes of artificial intelligence (AI) systems have been enabled
by supervised learning algorithms that require a large amount of human-labelled
data. Such data is costly to create, and it can be prohibitively expensive for
structured prediction tasks such as dependency
parsing~\citep{bohmova2003,brants2003}. Transfer learning~\citep{pan2010} is a
promising solution to facilitate the development of AI systems on a domain
without such data. In this work, we focus on a particular case of transfer
learning, namely cross-lingual learning, which seeks to transfer across
languages. We consider the setup where the target language is low-resource
having only unlabelled data, commonly referred to as unsupervised cross-lingual
transfer. This is an important problem because most world's languages are
low-resource~\citep{joshi2020}. Successful transfer from high-resource languages
enables language technologies development for these low-resource languages.

One recent method for unsupervised cross-lingual transfer is
PPTX~\citep{kurniawan2021}. It is developed for dependency parsing and allows
transfer from multiple source languages, which has been shown to be generally
superior to transferring from just a single language~\citep[\textit{inter
  alia}]{mcdonald2011,duong2015a,rahimi2019}. An advantage of PPTX is that, in
addition to not requiring any labelled data in the target language, it does not
require access to any data in the source languages either, which is useful if
the source data is private. All it needs is access to multiple, trained source
parsers. Despite its benefits, PPTX has only been applied to dependency parsing,
although in principle it should be extensible to other structured prediction
problems. More concerningly, we show in this work that PPTX generally
underperforms compared to a multi-source transfer baseline based on majority
voting.

\begin{figure}
  \centering
  \includegraphics[width=0.99\linewidth]{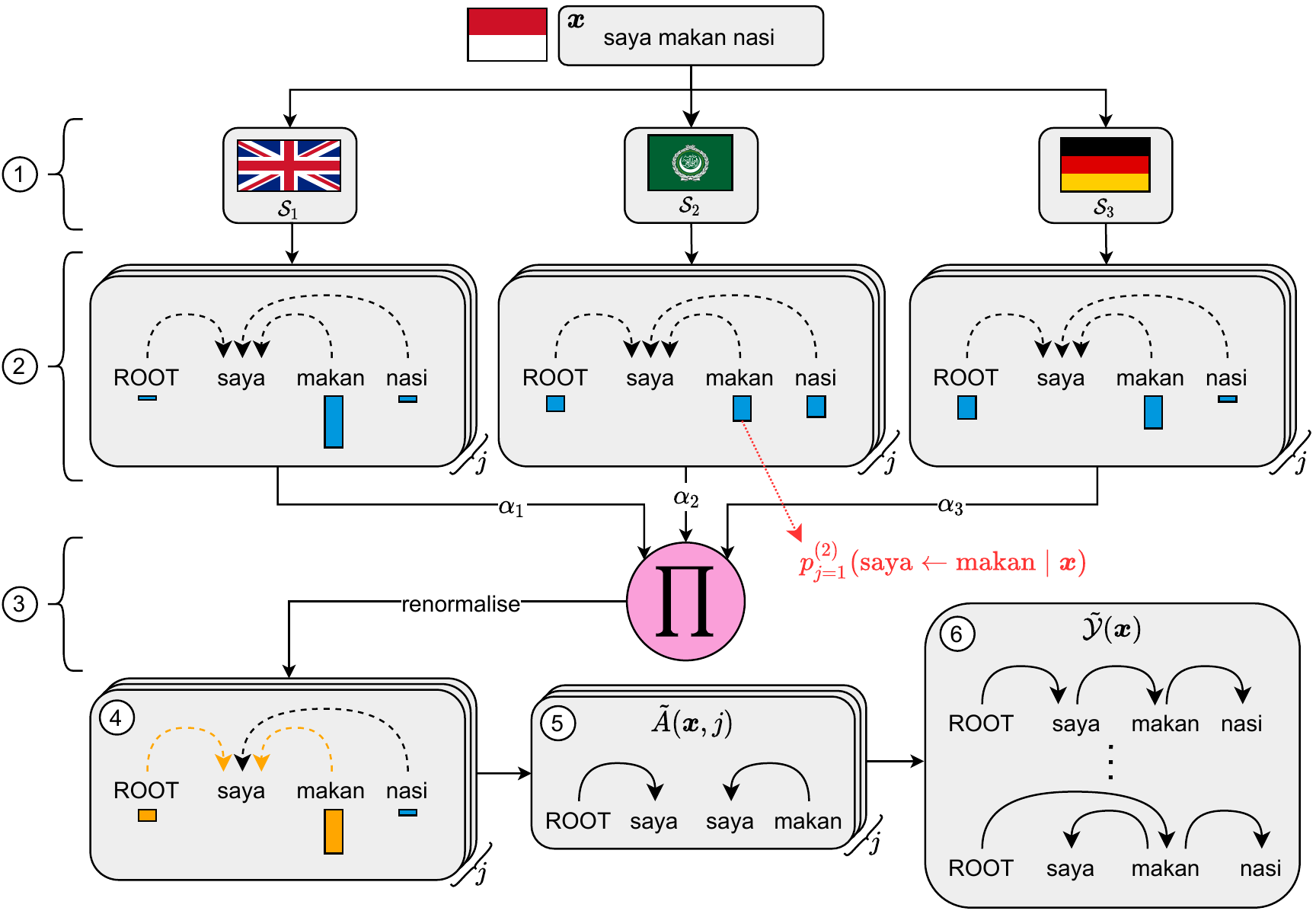}
  \caption{Illustration of \gmean{} for an input sentence \textit{saya makan nasi}
    (``I eat rice''). \Circled{1} A set of structured prediction models as
    inputs. \Circled{2} The models compute marginal probability distributions
    over substructures for each token $x_j$. \Circled{3} Logarithmic opinion
    pool of the distributions is computed. \Circled{4} Substructures are
    filtered based on some threshold. \Circled{5} High-probability substructures
    are obtained. \Circled{6} High-probability structures are obtained from the
    substructures as distant supervision.}\label{fig:ilustr}
\end{figure}

In this paper, we generalise and improve PPTX by reformulating it for structured
prediction problems. As with PPTX, this generalisation casts the unsupervised
transfer problem as a supervised learning task with distant supervision, where
the label of each sample in the target language is based on the structures predicted
by an ensemble of source models. Moreover, we propose the use of logarithmic
opinion pooling~\citep{heskes1998} to improve performance (see
\cref{fig:ilustr}). Unlike PPTX that performs simple union, the pooling
considers the output probabilities in aggregating the source model outputs to
obtain the structures used for distant supervision. We test \gmean{} on 18
languages from 5 language families and on two structured prediction tasks in
NLP: dependency parsing and part-of-speech tagging. We find that \gmean{}
generally outperforms both PPTX and the majority voting baseline, with absolute
accuracy gains of up to \SI{7}{\percent} on parsing and \SI{20}{\percent} on
tagging. Our analysis shows that the use of logarithmic opinion pooling results
in fewer predicted structures that are also more concentrated on the correct
ones.

In summary, our contributions in this paper are:
\begin{itemize}
  \item developing a generic unsupervised multi-source transfer method for
    structured prediction problems;
  \item leveraging logarithmic opinion pooling to take into account source
    model probabilities in the aggregation to produce the labels for distant
    supervision; and
  \item outperforming previous work in dependency parsing and
    part-of-speech tagging, especially in the context of a stronger,
    multi-source transfer baseline.
\end{itemize}

\section{Unsupervised Transfer as Supervised Learning}

Suppose we want to create a model for a low-resource language
that has only unlabelled data, but we only have access to a set of models
trained on other languages. This is an instance of cross-lingual transfer
learning. We cast this problem as a (distantly) supervised learning task with
the training objective
\begin{equation}
  \ell(\boldsymbol{\theta})
  =-\sum_{\boldsymbol{x}\in\mathcal{D}}
  \log\sum_{\boldsymbol{y}\in\tilde{\mathcal{Y}}(\boldsymbol{x})}
  p(\boldsymbol{y}\mid\boldsymbol{x};\boldsymbol{\theta})\label{eqn:loss}
\end{equation}
where $\boldsymbol{\theta}$ is the target model parameters, $\mathcal{D}$ is the
unlabelled target data, and $\tilde{\mathcal{Y}}(\boldsymbol{x})$ is a set of
distant supervision labels for an unlabelled input $\boldsymbol{x}=x_1x_2\cdots
x_n$. Thus, $\tilde{\mathcal{Y}}(\boldsymbol{x})$ contains supervision in the
form of one or more potentially ambiguous/uncertain labels. In single-source
transfer, $\tilde{\mathcal{Y}}(\boldsymbol{x})$ can be as simple as a singleton
containing the predicted label for $\boldsymbol{x}$ by the source model, in
which case this is related to self-training~\citep{mcclosky2006}. In our case,
however, this supervision is assumed to arise from an ensemble of models, each
is based on transfer from a different source language (see next section). The
parameters $\boldsymbol{\theta}$ can be initialised to the source model
parameters and regularised to this initialiser during training, in order to both
speed up training and encourage the parameters to stay near known good parameter
values. Overall, the objective becomes
$\ell'(\boldsymbol{\theta})=\ell(\boldsymbol{\theta})+\lambda\lVert
\boldsymbol{\theta}-\boldsymbol{\theta}_0 \rVert^2_2$ where
$\boldsymbol{\theta}_0$ is the source model parameters and $\lambda$ is a
hyperparameter controlling the regularisation strength.

\subsection{Supervision via Ensemble}

In multi-source transfer, the set $\tilde{\mathcal{Y}}(\boldsymbol{x})$ can be
obtained by an ensemble method applied to the source models.
PPTX~\citep{kurniawan2021} is one such method designed for arc-factored
dependency parsers. We generalise PPTX, making it applicable to any set of
source models that predict structured outputs that decompose into substructures
(of which a set of arc-factored dependency parsers is a special case).
For the rest of this paper, we assume that the source models are graphical
models over these structured outputs. Let $C(\boldsymbol{x},j)$ denote the set
of substructures associated with $x_j$ whose marginal probabilities form a
probability distribution:
\begin{equation}
  \sum_{c\in C(\boldsymbol{x},j)}p^{(k)}(c\mid\boldsymbol{x})=1
\end{equation}
for any source model $k$. For example, for dependency parsing,
$C(\boldsymbol{x},j)$ is the set of arcs whose dependent is $x_j$ (see
\cref{fig:ilustr} part \Circled{2}). The chart
$\tilde{\mathcal{Y}}(\boldsymbol{x})$ can then be obtained as follows. Define
$\tilde{A}_k(\boldsymbol{x},j)$ to be the set of substructures associated with
$x_j$ having high marginal probability under source model $k$. This set is
obtained by adding substructures $c\in C(\boldsymbol{x},j)$ in descending order
of their marginal probability until their cumulative probability exceeds a
threshold $\sigma$:
\begin{equation}
  \sum_{c\in C(\boldsymbol{x},j)}p^{(k)}(c\mid\boldsymbol{x})\geq\sigma\label{eqn:thresh}
\end{equation}
where $0\leq\sigma\leq 1$.
Therefore, $\tilde{A}_k(\boldsymbol{x},j)$ contains the substructures that cover
at least $\sigma$ probability mass of the output space under source model $k$.
Next, define
\begin{equation}
  \tilde{A}(\boldsymbol{x})
  =\bigcup_{k,j}\tilde{A}_k(\boldsymbol{x},j)\label{eqn:union}
\end{equation}
as the set of high probability substructures for $\boldsymbol{x}$ given by the
source models. The chart $\tilde{\mathcal{Y}}(\boldsymbol{x})$ is then defined as
the set of structures whose substructures are all in
$\tilde{A}(\boldsymbol{x})$. Formally,
\begin{equation}
  \tilde{\mathcal{Y}}(\boldsymbol{x})
  =\{\boldsymbol{y}\mid
  \boldsymbol{y}\in\mathcal{Y}(\boldsymbol{x})\wedge
  A(\boldsymbol{y})\subseteq\tilde{A}(\boldsymbol{x})
  \}
\end{equation}
where $\mathcal{Y}(\boldsymbol{x})$ is the output space of $\boldsymbol{x}$ and
$A(\boldsymbol{y})$ is the set of substructures in $\boldsymbol{y}$. To prevent
$\tilde{\mathcal{Y}}(\boldsymbol{x})$ from being empty, the 1-best structure
$\hat{\boldsymbol{y}}=\arg\max_{\boldsymbol{y}}p^{(k)}(\boldsymbol{y}\mid\boldsymbol{x})$
from each source model $k$ is also included in the chart, but they don't count
toward the probability threshold.

\subsection{Proposed Method}

Multilinguality is the key factor contributing to the success of
PPTX~\citep{kurniawan2021}. Therefore, optimising the method to leverage this
multilinguality provided by the source models is important. One potential
limitation of PPTX is the inclusion of substructures having relatively low
marginal probability under some source model because of the union in
\cref{eqn:union}. As an extreme illustration, consider a poor source model $k$
assigning uniform marginal probability to substructures in
$C(\boldsymbol{x},j)$. Most of these substructures will be included in
$\tilde{A}_k(\boldsymbol{x},j)$ and, subsequently, $\tilde{A}(\boldsymbol{x})$.
As a result, noisy structures may be included in
$\tilde{\mathcal{Y}}(\boldsymbol{x})$ which makes learning the correct structure
difficult.

Instead of computing the set of high-probability substructures from each source
model separately, a potentially better alternative is to aggregate the marginal
probabilities given by the source models and then compute the chart from the
resulting distribution. We propose to use logarithmic opinion
pooling~\citep{heskes1998} as the aggregation method. To obtain the chart
$\tilde{\mathcal{Y}}(\boldsymbol{x})$, first we compute the logarithmic opinion
pool of the source models' marginal probabilities. That is, for all
$j\in\{1,\ldots,n\}$, define
\begin{equation}
  \bar{p}_j(c\mid\boldsymbol{x})\propto
  \prod_k\left[ p^{(k)}(c\mid\boldsymbol{x}) \right]^{\alpha_k}\label{eqn:lop}
\end{equation}
where we normalise over the substructures $c\in C(\boldsymbol{x},j)$, and
$\alpha_k$ is a non-negative scalar weighting the contribution of source model
$k$ satisfying $\sum_k\alpha_k=1$. Thus, $\bar{p}_j$ gives the new probability
distribution over substructures in $C(\boldsymbol{x},j)$. Then, we compute the
set $\tilde{A}(\boldsymbol{x},j)$ using $\bar{p}_j$ in a similar fashion as
before: adding substructures $c\in C(\boldsymbol{x},j)$ in descending order by
their marginal probability given by $\bar{p}_j$ until their cumulative
probability exceeds $\sigma$. Lastly, we define
$\tilde{A}(\boldsymbol{x})=\bigcup_j\tilde{A}(\boldsymbol{x},j)$, and keep the
definition of $\tilde{\mathcal{Y}}(\boldsymbol{x})$ unchanged: the set of
structures induced by $\tilde{A}(\boldsymbol{x})$ plus the 1-best structures,
which is used as labels for training with the objective in \cref{eqn:loss}.
\cref{fig:ilustr} illustrates the process using dependency parsing as an
example.

\paragraph{Setting the Weight Factors}

\begin{figure}
  \centering
  \includegraphics[width=0.95\linewidth]{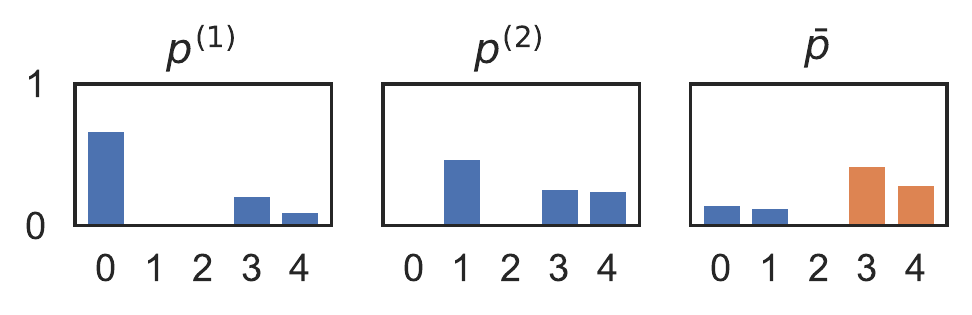}
  \caption{Logarithmic opinion pool with uniform weighting ($\bar{p}$) for two
    distributions $p^{(1)}$ and $p^{(2)}$. The opinion pool $\bar{p}$ assigns
    lower probabilities to substructures indexed by 0 and 1 than those indexed
    by 3 and 4 because $p^{(1)}$ and $p^{(2)}$ assign very low probability to
    either 0 or 1. Selected substructures in the context of \cref{eqn:thresh}
    with $\sigma=0.7$ are in orange.}\label{fig:lop}
\end{figure}

Finding an optimal value for $\alpha_k$ is possible if there is labelled
data~\citep{heskes1998}. However, we do not have labelled data in the target
language in our cross-lingual setup. There is some method to find
similar weighting scalars for cross-lingual transfer that may work in our
setup~\citep{wu2020a}, but they assume access to unlabelled source language data
and only marginally outperform uniform weighting. Therefore, unless stated
otherwise, we set $\alpha_k$ uniformly, reducing \cref{eqn:lop} to the
normalised geometric mean of the marginal distributions.

\paragraph{Motivation}

The motivation behind the proposed method is the observation that PPTX
obtains the high-probability substructures by applying the threshold in
\cref{eqn:thresh} for each source model separately \emph{before} they are
aggregated into a single set in \cref{eqn:union}. This means PPTX ``trusts'' all
source models equally regardless of their certainty about their predictions. In
contrast, \gmean{} takes into account the probabilities given by the source
models by applying the threshold \emph{after} aggregating the probabilities in
the logarithmic opinion pool in \cref{eqn:lop}. The opinion pool assigns more
probability mass to substructures to which all the source models assign a high
probability (see \cref{fig:lop}), and we hypothesise that such substructures are
more likely to be correct.

\subsection{Application to Structured Prediction}

The above method can be applied to structured prediction problems. Crucial
to the application is the definition of $C(\boldsymbol{x},j)$. Below, we present
two applications: arc-factored dependency parsing and sequence tagging.

\paragraph{Arc-Factored Dependency Parsing}

For dependency parsing, we can define $C(\boldsymbol{x},j)$ as the
set of dependency arcs having $x_j$ as dependent:
\begin{equation}
  C(\boldsymbol{x},j)
  =\{(i,j,l)\mid i\in\{0,1,\ldots,n\}\wedge i\neq j\wedge l\in L\}
\end{equation}
where $(i,j,l)$ denotes an arc from head $x_i$ to dependent $x_j$ with
dependency label $l$, $L$ denotes the set of dependency labels, and $x_0$ is a
special token whose dependent is the root of the sentence.\footnote{This
  formulation is widely used in graph-based dependency parsing, which dates back
  to the work of \citet{mcdonald2005}.} Since exactly one arc in
$C(\boldsymbol{x},j)$ exists in any possible dependency tree of
$\boldsymbol{x}$, the marginal probabilities of arcs in $C(\boldsymbol{x},j)$
form a probability distribution. The rest follows accordingly. The set
$\tilde{A}_k(\boldsymbol{x},j)$ becomes the set of arcs with $x_j$ as dependent
that have high marginal probability under source model $k$. The set
$\tilde{A}(\boldsymbol{x})$ becomes the set of high probability arcs for
the whole sentence. Lastly, the chart $\tilde{\mathcal{Y}}(\boldsymbol{x})$
contains all possible trees for $\boldsymbol{x}$ whose arcs are all in
$\tilde{A}(\boldsymbol{x})$. The predicted tree from each source parser is
included in the chart as well.

\paragraph{Sequence Tagging}

\begin{figure*}
  \centering
  \begin{subfigure}{0.9\textwidth}
    \includegraphics[width=\textwidth]{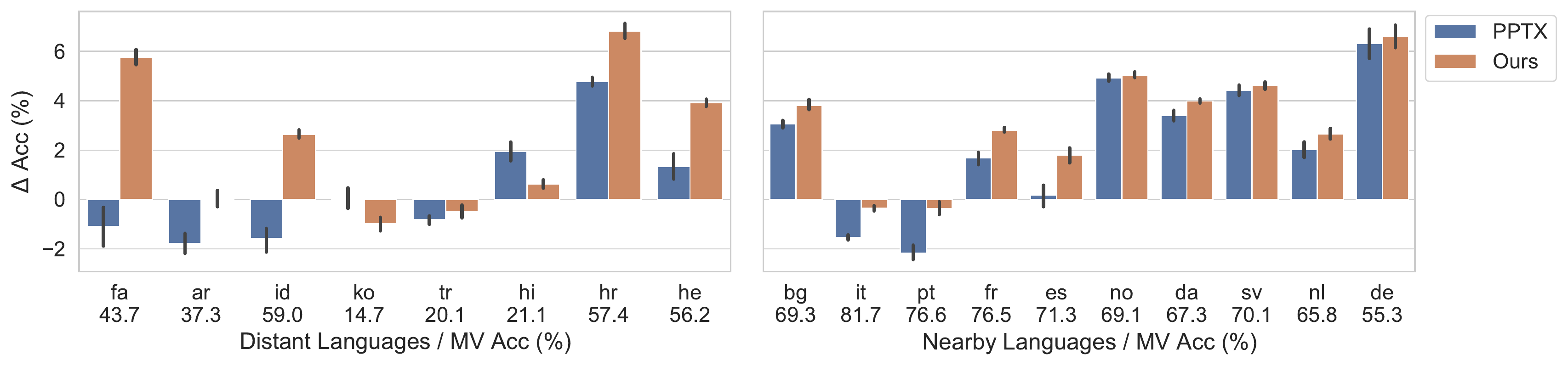}
    \caption{Dependency parsing}\label{fig:las-diff}
  \end{subfigure}
  \begin{subfigure}{0.9\textwidth}
    \includegraphics[width=\textwidth]{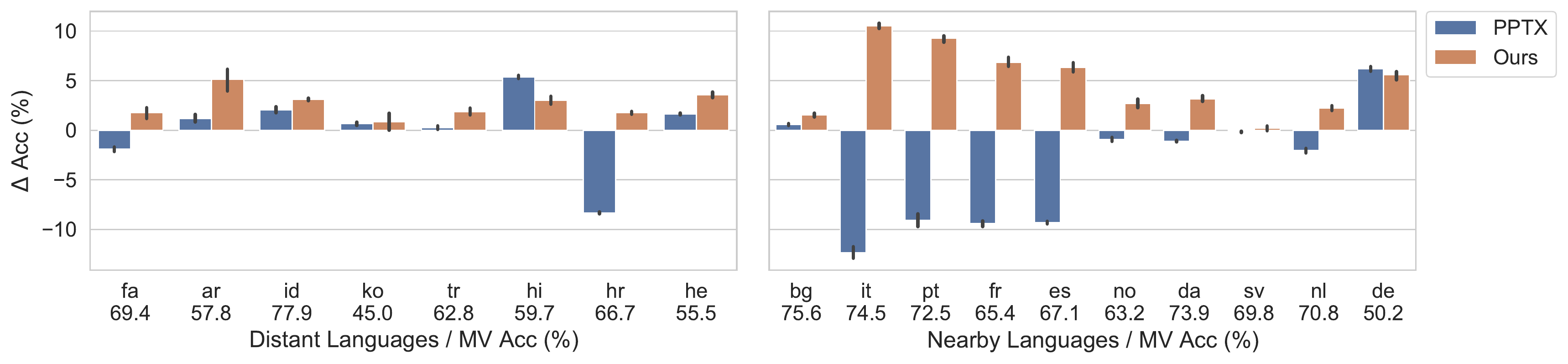}
    \caption{POS tagging}\label{fig:acc-diff}
  \end{subfigure}
  \caption{Performance difference of PPTX and \gmean{} against the majority
  voting baseline on dependency parsing and POS tagging. Numbers below the
  language label are the majority voting baseline performance, which corresponds
  to the zero value on the y-axis.}\label{fig:perf-diff}
\end{figure*}

In sequence tagging, the structured output is a sequence of tags, which
decomposes into consecutive tag pairs. Given a sequence of tags
$\boldsymbol{y}=y_1y_2\cdots y_n$ corresponding to the input
$\boldsymbol{x}$, its consecutive tag pairs are
$A(\boldsymbol{y})=\{(j,y_j,y_{j+1})\}_{j=1}^{n-1}$. We define
$C(\boldsymbol{x},j)$ as the set of possible tag pairs for $x_j$ and $x_{j+1}$:
\begin{equation}
  C(\boldsymbol{x},j)=\{(j,t,t')\mid(t,t')\in T\times T\}
\end{equation}
where $T$ is the set of tags. Note that any sequence of tags for
$\boldsymbol{x}$ has exactly one tag pair in $C(\boldsymbol{x},j)$ and thus,
the marginal probabilities of these tag pairs in $C(\boldsymbol{x},j)$ form a
probability distribution. With this definition, $\tilde{A}_k(\boldsymbol{x},j)$
becomes the set of tag pairs for $x_j$ and $x_{j+1}$ that have high marginal
probability under source model $k$, $\tilde{A}(\boldsymbol{x})$ becomes the set
of high probability tag pairs for $\boldsymbol{x}$ given by the source taggers,
and the chart $\tilde{\mathcal{Y}}(\boldsymbol{x})$ contains all possible tag
sequences for $\boldsymbol{x}$ whose consecutive tag pairs are all in
$\tilde{A}(\boldsymbol{x})$, plus the 1-best sequences from all the source
taggers.

\section{Experimental Setup}

\paragraph{Data and Evaluation}

We evaluate on dependency parsing and part-of-speech (POS) tagging. We use
Universal Dependencies v2.2~\citep{nivre2018} and test on 18 languages spanning
5 language families (see Supplementary Material). We divide the languages into
distant and nearby groups based on their distance to English~\citep{he2019a}. We
use the universal POS tags (UPOS) as labels for tagging. We exclude punctuation
from parsing evaluation, and report average performance across five random seeds
for PPTX and \gmean{} on both tasks. We include a PPTX baseline applied to
tagging hereinafter, even though it was originally developed for parsing. Our
evaluation metric is accuracy for both tasks. In dependency parsing, this metric
translates to the labelled attachment score (LAS), defined as the fraction of
correct labelled dependency relations. In POS tagging, accuracy is defined as
the fraction of correctly predicted POS tags. Both metrics are widely used by
previous work, and thus enable a fair comparison to ours.

\paragraph{Model Architecture}

For parsing, we use the same architecture as was used by
\citet{kurniawan2021}, which consists of embedding layers, a Transformer encoder
layer, and a biaffine output layer~\citep{dozat2017}. At test time, we run the
maximum spanning tree algorithm~\citep{chu1965,edmonds1967} to find the highest
scoring tree. For tagging, the same architecture is used but we replace the
output layer with a linear CRF layer. At test time, the Viterbi algorithm is
used to obtain the tag sequence with the highest score.

\paragraph{Source Selection}

We adopt a ``pragmatic'' approach where we include 5 high-resource
languages as sources: English, Arabic, Spanish, French, and
German~\citep{kurniawan2021}. These languages have been categorised as
``quintessential rich-resource languages'' due to the availability of massive
language datasets~\citep{joshi2020}. Other than English, all of these source
languages are in the set of target languages, so in that case, we exclude the
language from the sources. For example, if Arabic is the target language, then
we use only the other 4 languages as sources, thus the target language is
always unseen. To train the source models, we perform hyperparameter tuning on
English and use the values for training on the other source languages.
Generally, the source models achieve in-language performance comparable to
previous work~\citep[e.g.,][]{ahmad2019} with the exception of the Arabic parser
whose accuracy is noticeably lower, which we suspect is caused by the model
architecture optimised for transfer rather than in-language evaluation. However,
we argue that the lower performance indeed reflects a realistic application
scenario where some of the source models are expected to be of lower quality.
See Supplementary Material for details.

\paragraph{Baseline}

Our baseline is a majority voting ensemble (MV). For parsing, this is achieved
by scoring each possible arc by the number of source parsers that have it in
their predicted tree and then running the maximum spanning tree algorithm. For
tagging, we simply use the most commonly predicted tag for each input token. We
note that this baseline is more appropriate for multi-source transfer than the
direct transfer baseline used by \citet{kurniawan2021} which only uses a single
source language (English). We find MV is much stronger than direct transfer,
with accuracy gains of up to 15 points on both tasks.

\paragraph{Training}

We use the same setup as \citet{kurniawan2021} for parsing. We
include the gold universal POS tags as input to the parsers. We discard
sentences longer than 30 tokens to avoid memory issues and train for 5 epochs. We
tune the learning rate and $\lambda$ on the development set of Arabic, select
the values that give the highest accuracy, and use them for training on
all languages. For tagging, we set the length cut-off to 60 tokens and train
for 10 epochs. Similarly, we use only Arabic as the development language for
hyperparameter tuning, and use the best values for training on all languages.
For both tasks, we obtain cross-lingual word embeddings using an
offline transformation method~\citep{smith2017} applied to fastText pre-trained
word vectors~\citep{bojanowski2017}. We set the threshold $\sigma=0.95$
following \citet{kurniawan2021}. We set the parameters of the English source
model as $\boldsymbol{\theta}_0$. Further details can be found in Supplementary
Material.

\section{Results and Analysis}

\begin{table}\small
  \centering
  \sisetup{table-number-alignment=right}
  \begin{tabular}{@{}cS[table-format=1.1e1]S[table-format=1.4]S[table-format=1.1e1]S[table-format=3]@{}}
    \toprule
    \multirow{2}{*}{Language} & \multicolumn{2}{c}{Parsing} & \multicolumn{2}{c}{Tagging} \\
    \cmidrule(lr){2-3} \cmidrule(l){4-5}
       & {$n_P$ (in millions)} & {$\frac{n_O}{n_P}$ (\%)} & {$n_P$} & {$\frac{n_O}{n_P}$ (\%)} \\
    \midrule
    fa & 1.6e6                 & 0.0011                   & 6.5e5   & 3                        \\
    ko & 2.3e4                 & 0.021                    & 8.2e3   & 11                       \\
    hr & 2.0e5                 & 0.0019                   & 4.3e5   & 37                       \\
    it & 4.5                   & 0.069                    & 4.7e4   & 32                       \\
    es & 3.7e3                 & 0.0014                   & 2.4e6   & 110                      \\
    sv & 5.1                   & 0.12                     & 7.6e3   & 18                       \\
    \bottomrule
  \end{tabular}
  \caption{Median chart size of PPTX (column $n_P$), and median chart size of
    \gmean{} relative to PPTX (column $\frac{n_O}{n_P}$), where chart size is
    defined as the number of structures in
    $\tilde{\mathcal{Y}}(\boldsymbol{x})$.}\label{tbl:chartsz}
\end{table}

\begin{table*}\small
  \centering
  \sisetup{table-format = 2.0, table-auto-round}
  \begin{tabular}{@{}c*{6}{S}S[table-format=2.1]*{6}{S}S[table-format=2.1]@{}}
    \toprule
    \multirow{3}{*}{Target Language} & \multicolumn{7}{c}{Parsing} & \multicolumn{7}{c}{Tagging} \\
    \cmidrule(lr){2-8} \cmidrule(l){9-15}
    & \multicolumn{2}{c}{PPTX} & \multicolumn{2}{c}{\gmn{}} & \multicolumn{3}{c}{$\Delta$} & \multicolumn{2}{c}{PPTX} & \multicolumn{2}{c}{\gmn{}} & \multicolumn{3}{c}{$\Delta$} \\
    \cmidrule(lr){2-3} \cmidrule(lr){4-5} \cmidrule(lr){6-8} \cmidrule(lr){9-10} \cmidrule(lr){11-12} \cmidrule(l){13-15}
       & {P}  & {R}  & {P}  & {R}  & {P}  & {R}  & {Acc} & {P}  & {R}  & {P}  & {R}  & {P}  & {R}  & {Acc} \\
    \cmidrule(r){1-8} \cmidrule(l){9-15}
    fa & 10.4 & 89.6 & 17.4 & 94.7 & 7.0  & 5.1  & 6.9   & 20.7 & 79.5 & 25.9 & 75.2 & 5.2  & 4.3  & 3.7   \\
    ko & 0.4  & 64.9 & 1.8  & 77.4 & 1.4  & 12.5 & -1.0  & 8.4  & 49.5 & 3.5  & 44.8 & -4.9 & -4.7 & 0.2   \\
    hr & 10.1 & 95.6 & 20.4 & 98.3 & 10.3 & 2.7  & 2.1   & 14.2 & 76.5 & 15.8 & 76.7 & 1.6  & 0.2  & 10.1  \\
    it & 10.4 & 99.0 & 25.3 & 99.6 & 14.9 & 0.6  & 1.2   & 20.0 & 90.7 & 24.3 & 93.1 & 4.3  & 2.4  & 22.9  \\
    es & 10.6 & 96.3 & 20.3 & 97.5 & 9.7  & 1.2  & 1.6   & 17.6 & 82.4 & 16.3 & 86.5 & -1.3 & 4.1  & 15.7  \\
    sv & 12.9 & 97.2 & 20.8 & 98.4 & 7.9  & 1.2  & 0.2   & 20.6 & 83.8 & 28.3 & 81.0 & 7.7  & -2.8 & 0.4   \\
    \bottomrule
  \end{tabular}
  \caption{Precision (P) and recall (R) of charts produced by PPTX and \gmean{}
    in dependency parsing and POS tagging. Numbers are rounded to the nearest
    integer. Column $\Delta$ is the difference between \gmean{} and PPTX
    (positive means \gmean{} is higher). $\Delta$ over the accuracy results for
    both tasks are included for completeness, and correspond to the bar height
    difference of the two methods in \cref{fig:perf-diff}.}\label{tbl:chartpr}
\end{table*}

\cref{fig:las-diff} shows the accuracy difference of PPTX and \gmean{} against MV
on parsing. We see that PPTX does not consistently outperform MV,
substantially underperforming on 6 languages.\footnote{Persian, Arabic,
  Indonesian, Turkish, Italian, and Portuguese.} On the other hand, \gmean{}
outperforms not only PPTX but also MV on most languages. \cref{fig:acc-diff}
shows the corresponding results on POS tagging which is particularly convincing. We
see that PPTX often underperforms, with up to \SI{10}{\percent} drop in accuracy
compared to MV. In contrast, \gmean{} consistently outperforms MV with up to
\SI{10}{\percent} accuracy improvement. These results suggest that PPTX may not
improve over a simple majority voting ensemble, and \gmean{} is the
superior alternative.
In addition, \gmean{} shows higher
improvement against MV on nearby than distant languages, which is unsurprising
because our pragmatic selection of source languages is dominated by languages
in the nearby group.

From the figure, we also see that on Portuguese and Italian, \gmean{} slightly
underperforms compared to MV on parsing, but outperforms MV considerably on
tagging. We hypothesise that this disparity is caused by the variability of the
source models quality. On tagging, the direct transfer performance of 3 out of 5
source taggers is relatively poor on Portuguese and Italian, making it more
likely for MV to predict wrongly as the good taggers are outvoted. In contrast,
on parsing, Arabic is the only source parser that has very poor transfer. The
other source parsers achieve comparably good direct transfer performance so MV
already performs well.

\subsection{Chart Size Analysis}

To understand the differences between PPTX and \gmean{} better, we look at the
chart produced by the two methods. Specifically, we compare the size of the
chart $\tilde{\mathcal{Y}}(\boldsymbol{x})$ produced by PPTX and \gmean{}, in
terms of the number of structures in it. We take the median of
this size over all unlabelled sentences in the training set of each target
language and compare the results. \cref{tbl:chartsz} reports the median chart
size of PPTX, and the median chart size of \gmean{} relative to PPTX for both
parsing and tagging on 6 representative languages (the trend for other languages is
similar).  We find that for parsing, the size of \gmean{}'s chart is
much smaller than \SI{1}{\percent} of the size of PPTX chart for all target
languages.\footnote{Except for Turkish, where this number is \SI{3}{\percent},
  which is still very small.} This finding shows that \gmean{}'s charts are much
more compact than those of PPTX. Thus, it may explain the improvement of
\gmean{} over PPTX because smaller charts may be more likely to concentrate on
trees that have many correct arcs, making it easier for the model to learn
correctly (we explore this further in the next section). For POS tagging, we
find the same trend as in dependency parsing where \gmean{}'s charts are
smaller, but to a lesser extent, presumably because the typical output space of
tagging is several orders of magnitude smaller than that of parsing.
Occasionally, \gmean{}'s chart is larger than that of PPTX, although \gmean{}
outperforms PPTX substantially (French and Spanish). We speculate that this is
because most of the source taggers are very confident but on different
substructures, so only a handful of substructures are selected by PPTX after
applying the threshold in \cref{eqn:thresh}, making the chart small. Meanwhile,
the logarithmic opinion pool is less confident as it corresponds to the
(geometric) mean of the distributions, so more substructures are selected,
making the chart larger.

\subsection{Chart Quality Analysis}

Continuing the previous analysis, we check if the smaller charts of \gmean{}
indeed concentrate more on the correct structures than those of PPTX. To measure
this, we define the notion of precision and recall of the chart
$\tilde{\mathcal{Y}}(\boldsymbol{x})$. We define precision as the fraction of
correct substructures in $\tilde{\mathcal{Y}}(\boldsymbol{x})$ and recall as the
fraction of gold substructures that occur in any structure in
$\tilde{\mathcal{Y}}(\boldsymbol{x})$. Formally,
\begin{equation}
  \mathrm{P}(\tilde{\mathcal{Y}}(\boldsymbol{x}))
  =\frac{
    \sum_{(\boldsymbol{x},\boldsymbol{y}^*)}
    \sum_{\boldsymbol{y}\in\tilde{\mathcal{Y}}(\boldsymbol{x})}
    |A(\boldsymbol{y})\cap A(\boldsymbol{y}^*)|
  }{
    \sum_{\boldsymbol{x}}
    \sum_{\boldsymbol{y}\in\tilde{\mathcal{Y}}(\boldsymbol{x})}
    |A(\boldsymbol{y})|
  }
\end{equation}
and
\begin{equation}
  \mathrm{R}(\tilde{\mathcal{Y}}(\boldsymbol{x}))
  =\frac{
    \sum_{(\boldsymbol{x},\boldsymbol{y}^*)}
    \sum_{a\in A(\boldsymbol{y}^*)}\mathrm{I}(a,\tilde{\mathcal{Y}}(\boldsymbol{x}))
  }{
    \sum_{\boldsymbol{y}^*}|A(\boldsymbol{y}^*)|
  }
\end{equation}
where
\begin{equation}
  \mathrm{I}(a,\tilde{\mathcal{Y}}(\boldsymbol{x}))
  =\begin{cases}
    1 & \text{if }\boldsymbol{y}\in\tilde{\mathcal{Y}}(\boldsymbol{x})\text{ s.t. }a\in A(\boldsymbol{y}) \\
    0 & \text{otherwise}
  \end{cases}
\end{equation}
and $\boldsymbol{y}^*$ denotes the gold structure for input $\boldsymbol{x}$. A
good chart must have high precision and recall. In particular, if
$\tilde{\mathcal{Y}}(\boldsymbol{x})$ is a singleton containing the gold
structure, then both precision and recall will be \SI{100}{\percent}.

\cref{tbl:chartpr} reports the precision and recall of the charts produced by
PPTX and \gmean{} for both tasks, as well as the performance differences, for
the same 6 languages as before (the trend for other languages is similar). We
observe that with \gmean{} in parsing, both precision and recall consistently
improve over PPTX, suggesting that the charts indeed contain more correct arcs.
However, higher precision and recall do not guarantee performance improvement,
as shown by Korean where both precision and recall improve with \gmean{} but its
performance is lower than PPTX.\footnote{The only other language where this
  happens is Hindi.} We suspect that this is caused by the unusually low
precision even with \gmean{}, indicating that the chart is very noisy. For POS
tagging, the result is less obvious, but we find that generally \gmean{}
improves chart precision, but often sacrificing chart recall. For Spanish,
precision decreases with \gmean{}, and only recall improves.\footnote{The only
  other language where this happens is French.} An interesting case is again
Korean where both precision and recall worsen, probably because of very poor
source taggers performance on the language. Overall, \gmean{} generally improves
the chart quality in terms of either precision or recall, but to a lesser
extent, which again may be attributed to the smaller output space compared with
parsing.

\subsection{Effect of Opinion Pool Distance to True Distribution}

\begin{figure}
  \centering
  \begin{subfigure}{0.49\linewidth}
    \includegraphics[width=\linewidth]{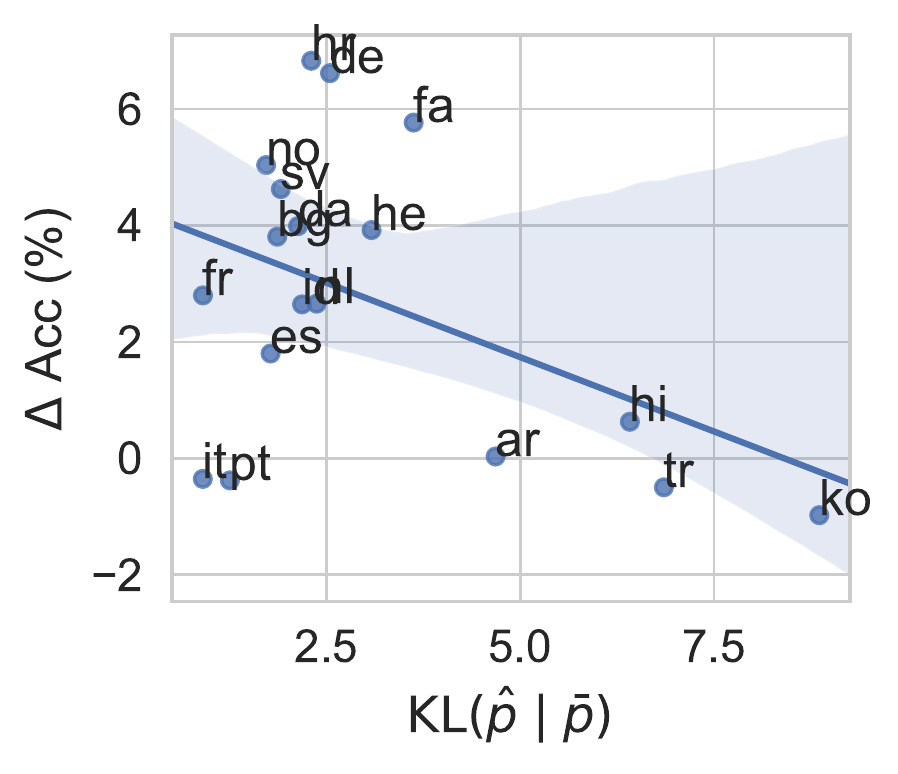}
    \caption{Dependency parsing}\label{fig:ea-vs-dlas}
  \end{subfigure}
  \begin{subfigure}{0.49\linewidth}
    \includegraphics[width=\linewidth]{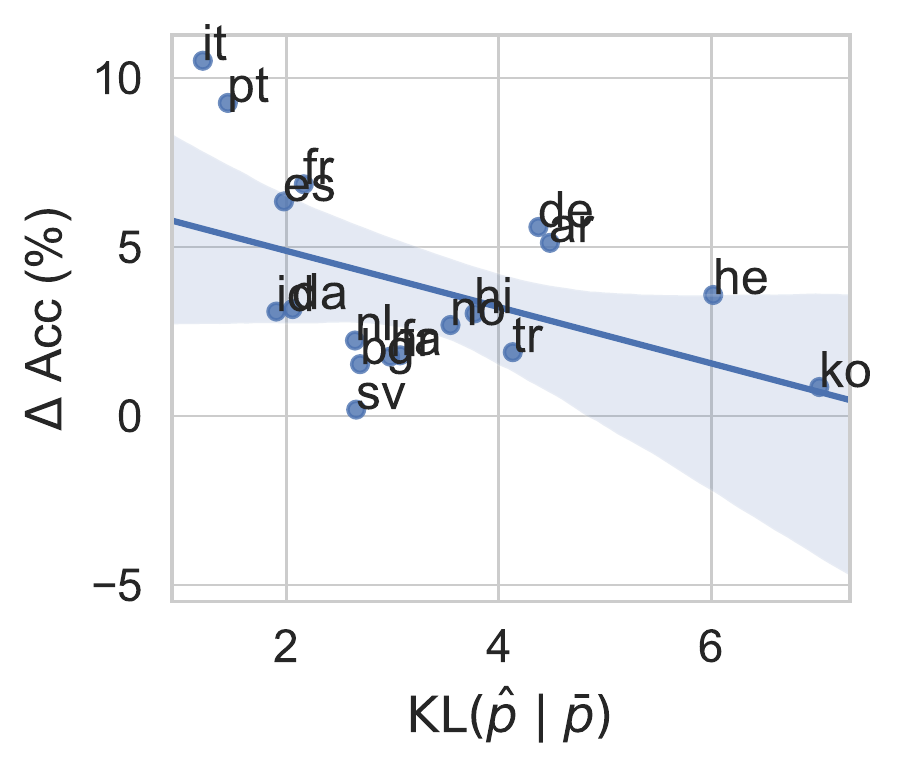}
    \caption{POS tagging}\label{fig:ea-vs-dacc}
  \end{subfigure}
  \caption{Relationship between $\mathrm{KL}(\hat{p}\mid\bar{p})$ and the accuracy
    difference of \gmean{} and MV, where $\hat{p}$ and $\bar{p}$ denote the
    empirical true distribution and the opinion pool distribution respectively.
    Shaded area is \SI{95}{\percent} confidence interval computed via
    bootstrapping.}\label{fig:ea-vs-dperf}
\end{figure}

We explore whether there is a relationship between (a) how distant the opinion
pool is to the true distribution over substructures and (b) the performance
improvement of \gmean{} against majority voting. Intuitively, the closer the
opinion pool is to the true distribution, the higher its absolute performance
would be. However, it is unclear whether this translates into an advantage over
majority voting. This is important because if such relationship exists, then it
may be worthwhile spending some effort on optimising the opinion pool. To this
end, we measure the distance between the true distribution and the opinion pool
by computing the Kullback-Leibler divergence (KL)
\begin{equation}
  \mathrm{KL}(\hat{p}\mid\bar{p})
  =\frac{1}{n(\mathcal{D})}\sum_{\boldsymbol{x}\in\mathcal{D}}\sum_{j=1}^{|\boldsymbol{x}|}\mathrm{KL}(\hat{p}_j\mid\bar{p}_j)\label{eqn:lop-dist}
\end{equation}
where
\begin{equation}
  \mathrm{KL}(\hat{p}_j\mid\bar{p}_j)
  =-\sum_{c\in C(\boldsymbol{x},j)}\hat{p}_j(c\mid\boldsymbol{x})\log
  \left[ \frac{\bar{p}_j(c\mid\boldsymbol{x})}{\hat{p}_j(c\mid\boldsymbol{x})} \right],
\end{equation}
$n(\mathcal{D})$ is the total number of tokens of all input sentences in $\mathcal{D}$, $\hat{p}_j$
is the (empirical) true distribution of substructures in $C(\boldsymbol{x},j)$, and
$\bar{p}_j$ is the logarithmic opinion pool distribution defined in
\cref{eqn:lop}. Note that $\hat{p}_j$ is a one-hot distribution so
$\mathrm{KL}(\hat{p}_j\mid\bar{p}_j)$ reduces to the negative log likelihood of
the labelled data under the opinion pool. We compute the KL divergence on the
training set of both parsing and tagging and display the regression plots in
\cref{fig:ea-vs-dperf}. We see a medium correlation between opinion
pool distance and performance gain against majority voting, with $r=-0.45$ for both
parsing and tagging ($p$-value is \num{0.06} for both). However, there
is substantial variance, especially in the right half figure of
parsing, caused by the lack of languages in that region of the plot.
Nonetheless, the plots suggest that there is indeed a positive relationship
between how close the opinion pool is to the true distribution and the
performance gain of \gmean{} compared with majority voting.

There are ways to obtain an opinion pool that is closer to the true
distribution. One way is to leverage a small amount of labelled data in
the target language to estimate the weight factors $\alpha_k$, which can be
done by optimising \cref{eqn:lop-dist}. This option is suitable if such
labelled data is available or can be obtained cheaply. There is some evidence
that 50 samples are enough to estimate similar weight factors in a linear
opinion pool~\citep{hu2021}, which may also apply to our setup. If we have
the freedom to choose the source languages, another method is to select them
carefully so they are both reasonably close to the target language and also
diverse. This is because \cref{eqn:lop-dist} can be expressed as the difference
between two terms, respectively corresponding to how distant the source
models' output distributions are to the target's true distribution
(\emph{error}) and how distant they are to each other
(\emph{diversity})~\citep{heskes1998}. Having the source languages reasonably
close to the target language and also diverse means reducing the first and
increasing the second term respectively, moving the opinion pool closer to
the true distribution. That said, when the source languages are close to the
target language, the source models may already be good for direct transfer so
\gmean{} may not give meaningful improvement over majority voting.

\subsection{Learning the Opinion Pool Weight Factors}

\begin{table}\small
  \centering
  \sisetup{table-format=2.1, table-auto-round}
  \begin{tabular}{@{}lSS@{}}
    \toprule
                       & {Parsing} & {Tagging} \\
    \midrule
    MV                 & 56.25     & 65.44     \\
    Uniform $\alpha_k$ & 58.97     & 69.30     \\
    Learned $\alpha_k$ & 59.35     & 69.96     \\
    \bottomrule
  \end{tabular}
  \caption{Parsing and tagging performance of MV and \gmean{} with uniform and
    learned weight factors $\alpha_k$ for the logarithmic opinion pool, averaged
    over 18 languages.}\label{tbl:lopw}
\end{table}

Motivated by the previous findings, we deviate from our unsupervised
setup by learning the weight factors $\alpha_k$ using a tiny amount of
labelled target data. Concretely, we randomly sample 50 sentences from the
training set of each target language and learn $\alpha_k$ that minimises
\cref{eqn:lop-dist} for all source model $k$. We then use the learned weights to
obtain the opinion pool as defined in \cref{eqn:lop} (see Supplementary Material
for further details). \cref{tbl:lopw} shows the results on parsing and tagging,
averaged over the target languages. We observe that by using the learned weight
factors, \gmean{} slightly improves over the version using uniform weights,
suggesting that \gmean{} can readily leverage labelled target data if it is
available. On the other hand, the fact that the improvement is only modest also
reaffirms that uniform weighting is a strong baseline.

\section{Related Work}

A straightforward method of multi-source transfer is training a model on the
concatenation of datasets from the source languages. This approach was used by
\citet{mcdonald2011} for dependency parsing. They find that this method yields a
strong performance gain compared with single-source transfer. More recent work
by \citet{guo2016} proposed to learn multilingual representations from the
concatenation of source language data and use them to train a neural dependency
parser. Another method is language adversarial training, used by
\citet{chen2019} for various NLP tasks including named-entity recognition, which
is another structured prediction problem. Despite their success, multi-source
unsupervised cross-lingual transfer methods typically assume access to the
source language data, which is not always feasible.

There is recent work suitable in this source-free setup. \citet{rahimi2019}
proposed a method based on truth inference to model label confusion in
multi-source transfer of named-entity recognisers. However, extending their
method to other structured prediction problems such as dependency parsing is not
straightfoward. \citet{wu2020a} used teacher-student learning to transfer from a
set of source models as ``teachers'' to a target model as ``student'' for
named-entity recognition. The method resembles knowledge distillation where the
student model is trained to predict soft labels from the teachers, in this case
given as a mixture of output distributions. They proposed a method to weight the
source models assuming access to unlabelled source data is possible.
\citet{hu2021} argued that in many cases, a small amount of labelled data is
available in the target language and proposed an attention-based method to
weight the source models leveraging such labelled data for structured
prediction. Their best method weights the source models at the substructure
level, which can be costly to run.

Our work builds upon the work of \citet{kurniawan2021} who proposed a method
based on self-training for unsupervised cross-lingual dependency parsing. Their
multi-source method builds a chart for every unlabelled sample in the target
language by combining high probability trees from the source parsers. In this
work, we generalise their method to structured prediction problems and propose a
modification to improve the quality of the generated charts.

\section{Conclusions}

In this paper, we (1) generalise previous methods for cross-lingual unsupervised
transfer without source data to structured prediction problems and (2) propose a
new aggregation technique which can better handle mixed-quality input
distributions. Experiments across two structured prediction tasks and 18
languages show that, unlike previous work, \gmean{} generally outperforms a
strong multi-source transfer baseline. Our analyses suggest that \gmean{}
produces distant supervision of better quality than that of the previous
methods. Our work potentially generalises beyond language transfer to (a)
structured prediction tasks beyond NLP and (b) transfer across other types of
domains (e.g., genres), a direction we aim to explore in future work.

\section*{Acknowledgements}

A graduate research scholarship is provided by Melbourne School of Engineering
to Kemal Kurniawan.

\nocite{*}
\bibliography{aaai22,ud}

\appendix

\clearpage

\section{Supplementary Material}

\subsection{Evaluation Languages}

\cref{tbl:lang} lists the languages we use in our evaluation, along with their
family, subgroup (if the language is Indo-European), and selected treebanks in
Universal Dependencies v2.2. This selection follows \citet{kurniawan2021} to
enable a fair comparison.

\begin{table}[!h]\small
  \centering
  \begin{tabular}{@{}llll@{}}
    \toprule
    Language   & Code & Family       & UD Treebanks       \\
    \midrule
    \multicolumn{4}{c}{\textit{Distant}}                  \\
    \addlinespace
    Persian    & fa   & IE.Iranian   & Seraji             \\
    Arabic     & ar   & Afro-Asiatic & PADT               \\
    Indonesian & id   & Austronesian & GSD                \\
    Korean     & ko   & Koreanic     & GSD, Kaist         \\
    Turkish    & tr   & Turkic       & IMST               \\
    Hindi      & hi   & IE.Indic     & HDTB               \\
    Croatian   & hr   & IE.Slavic    & SET                \\
    Hebrew     & he   & Afro-Asiatic & HTB                \\
    \midrule
    \multicolumn{4}{c}{\textit{Nearby}}                   \\
    \addlinespace
    Bulgarian  & bg   & IE.Slavic    & BTB                \\
    Italian    & it   & IE.Romance   & ISDT               \\
    Portuguese & pt   & IE.Romance   & Bosque, GSD        \\
    French     & fr   & IE.Romance   & GSD                \\
    Spanish    & es   & IE.Romance   & GSD, AnCora        \\
    Norwegian  & no   & IE.Germanic  & Bokmaal, Nynorsk   \\
    Danish     & da   & IE.Germanic  & DDT                \\
    Swedish    & sv   & IE.Germanic  & Talbanken          \\
    Dutch      & nl   & IE.Germanic  & Alpino, LassySmall \\
    German     & de   & IE.Germanic  & GSD                \\
    \bottomrule
  \end{tabular}
  \caption{List of languages in our evaluation, grouped into distant and nearby
    languages based on their distance to
    English~\citep{he2019a}. IE stands for Indo-European.}\label{tbl:lang}
\end{table}

\subsection{Source Models Performance}

\cref{tbl:src-las-acc} reports the performance of our source parsers and
taggers. We also report the performance numbers of previous work, copied from
their respective papers, to serve as reference.

\begin{table}[!h]\small
  \centering
  \sisetup{table-number-alignment=right}
  \begin{tabular}{@{}l*{5}{S}@{}}
    \toprule
                   & {en} & {ar} & {es} & {fr} & {de}           \\
    \midrule
    Parsing        & 86.9 & 76.9 & 90.0 & 89.1 & 82.1           \\
    Tagging        & 94.5 & 95.4 & 96.5 & 96.5 & 92.1           \\
    \midrule
    \multicolumn{6}{c}{\textit{Previous work (reference only)}} \\
    \addlinespace
    LSTM parser    & 88.3 & 81.8 & 90.8 & 89.1 & 83.7           \\
    Stanza tagger* & 95.4 & 94.9 & 96.7 & 97.3 & 94.1           \\
    \bottomrule
  \end{tabular}
  \caption{Parsing and tagging accuracy of the source models. We copy numbers of
    the LSTM parser~\citep{ahmad2019} and Stanza tagger~\citep{qi2018} from
    their respective papers to serve as reference only. * indicates that the
    numbers are not directly comparable to ours because of the difference in the
    evaluation setup.}\label{tbl:src-las-acc}
\end{table}

\subsection{Additional Experiment Details}

We implement \gmean{} using Python v3.7, PyTorch v1.4~\citep{paszke2019}, and
PyTorch-Struct~\citep{rush2020}. We run our experiments with Sacred
v0.8.2~\citep{greff2017a}, which also sets the random seeds. Experiments are run
on NVIDIA GeForce GTX TITAN X with CUDA 10.1 and GPU memory of 11 MiB. CPU model
is Intel(R) Xeon(R) CPU E5-2687W v3 @ 3.10GHz with Ubuntu 16.04 as the operating
system.

\subsection{Hyperparameters}

\begin{table}[!h]\small
  \centering
  \begin{tabular}{@{}llll@{}}
    \toprule
    Task                     & Method                  & Hyperparameter Dist.        & Best Value                  \\
    \midrule
    \multirow{4}{*}{Parsing} & \multirow{2}{*}{PPTX}   & $\log\eta\sim U(-6,-3)$     & $\eta=8.5\times 10^{-5}$    \\
                             &                         & $\log\lambda\sim U(-4,1)$   & $\lambda=2.8\times 10^{-5}$ \\
                             & \multirow{2}{*}{\gmn{}} & $\log\eta\sim U(-6.5,-3.5)$ & $\eta=9.4\times 10^{-5}$    \\
                             &                         & $\log\lambda\sim U(-4,1)$   & $\lambda=1.6\times 10^{-4}$ \\
    \addlinespace
    \multirow{4}{*}{Tagging} & \multirow{2}{*}{PPTX}   & $\log\eta\sim U(-6,-4)$     & $\eta=5.9\times 10^{-5}$    \\
                             &                         & $\log\lambda\sim U(-4,1)$   & $\lambda=0.1$               \\
                             & \multirow{2}{*}{\gmn{}} & $\log\eta\sim U(-6.5,-3.5)$ & $\eta=2.6\times 10^{-4}$    \\
                             &                         & $\log\lambda\sim U(-4,1)$   & $\lambda=4.7\times 10^{-3}$ \\
    \bottomrule
  \end{tabular}
  \caption{Distributions of hyperparameters we use for tuning on Arabic with
    random search and the best values found. All logarithms are of base
    10.}\label{tbl:hyperparam}
\end{table}

\begin{table}[!h]
  \centering
  \begin{tabular}{@{}lr@{}}
    \toprule
    Hyperparameter                    & Value \\
    \midrule
    Word embedding size               & 300   \\
    Word dropout                      & 0.2   \\
    $d_{\text{key}},d_{\text{value}}$ & 64    \\
    $d_{\text{ff}}$                   & 512   \\
    $n_{\text{head}}$                 & 8     \\
    $n_{\text{layer}}$                & 6     \\
    Batch size                        & 80    \\
    \midrule
    \multicolumn{2}{c}{\textit{Parsing-only}} \\
    \addlinespace
    POS tag embedding size            & 50    \\
    Output embedding dropout          & 0.2   \\
    $d_{\text{arc}}$                  & 512   \\
    $d_{\text{deptype}}$              & 128   \\
    \bottomrule
  \end{tabular}
  \caption{List of hyperparameter values used in our parsers and taggers.
    $d_{\text{key}},d_{\text{value}}$: size of key and value vector in the
    Transformer encoder. $d_{\text{ff}}$: size of feedforward network hidden layer in
    the Transformer encoder. $n_{\text{head}}$: number of heads in the
    Transformer encoder. $n_{\text{layer}}$: number of layers in the Transformer
    encoder. $d_{\text{arc}},d_{\text{deptype}}$: size of feedforward network
    output layer corresponding to arcs and dependency types in the biaffine
    output layer of parsers.}\label{tbl:other-hyperparam}
\end{table}

We tune learning rate $\eta$ and $\lambda$ using random search.
\cref{tbl:hyperparam} shows the distributions of each hyperparameter we use, and
the best values we find. We sample 20 values from the distribution and pick the
values that yield the best accuracy on the Arabic development set. We follow
\citet{kurniawan2021} for other hyperparameters, whose values are reported in
\cref{tbl:other-hyperparam}.

\subsection{Learning the Opinion Pool Weight Factors}

We learn the factors $\alpha_k$ weighting the contribution of source model $k$
in the logarithmic opinion pool by minimising \cref{eqn:lop-dist} with respect
to $\alpha_k$. The minimisation is done on 50 randomly sampled sentences from
the target language's training set using gradient descent. We set the initial
learning rate to \num{0.1} and reduce it at every epoch by a factor of
\num{0.9}. We initialise the weight factors uniformly at the start and run the
training until convergence. After the weight factors are learned, we use and fix
them for all subsequent experiments. We proceed with hyperparameter tuning on
Arabic using the same procedure as the version with uniform weights. For both
tasks, we tune $\eta$ and $\lambda$ with random search (20 runs), drawing from
$\log_{10}\eta\sim U(-6,-3)$ and $\log_{10}\lambda\sim U(-4,1)$ respectively.
For parsing, the best values are $\eta=9.1\times 10^{-5}$ and $\lambda=5.1\times
10^{-4}$. For tagging, they are $\eta=4.7\times 10^{-4}$ and $\lambda=0.062$.
These values are then used for the other languages. Lastly, we report the
average accuracy over the languages in \cref{tbl:lopw}.

\subsection{Full Experiment Results}

We report in \cref{tbl:full-results} the full results of MV, PPTX, and \gmean{}
(with both uniform and learned weight factors $\alpha_k$) on both dependency
parsing and POS tagging, averaged over 5 runs.

\begin{table*}\small
  \begin{subtable}{1\textwidth}
    \centering
    \sisetup{table-number-alignment=right, separate-uncertainty, table-format=2.1(1)}
    \begin{tabular}{@{}cS[table-format=2.1]*{3}{S}S[table-format=2.1]*{3}{S}@{}}
      \toprule
      \multirow{2}{*}{Language} & \multicolumn{4}{c}{Parsing} & \multicolumn{4}{c}{Tagging} \\
      \cmidrule(lr){2-5} \cmidrule(l){6-9}
         & {MV} & {PPTX}   & {\gmn{}} & {\gmn{}, learned $\alpha_k$} & {MV} & {PPTX}  & {\gmn{}} & {\gmn{}, learned $\alpha_k$} \\
      \midrule
      \multicolumn{9}{c}{\textit{Distant}}                                                                                      \\
      \addlinespace
      fa & 43.7 & 42.5(11) & 49.4(5)  & 48.9(3)                      & 69.2 & 67.5(2) & 71.2(6)  & 72.3(5)                      \\
      ar & 37.6 & 36.4(6)  & 38.9(5)  & 38.6(5)                      & 58.5 & 59.0(4) & 62.5(11) & 63.0(17)                     \\
      id & 56.8 & 54.8(8)  & 59.0(3)  & 59.1(1)                      & 77.5 & 79.6(4) & 81.0(2)  & 80.6(8)                      \\
      ko & 13.7 & 13.6(4)  & 12.8(3)  & 13.6(2)                      & 44.1 & 44.0(3) & 44.2(11) & 43.4(17)                     \\
      tr & 20.8 & 19.9(2)  & 20.2(4)  & 21.2(2)                      & 62.8 & 62.8(3) & 64.2(3)  & 64.3(2)                      \\
      hi & 21.9 & 23.9(5)  & 22.7(2)  & 27.2(2)                      & 59.9 & 65.6(2) & 63.2(6)  & 65.5(9)                      \\
      hr & 57.1 & 60.7(3)  & 62.9(4)  & 62.8(3)                      & 67.2 & 58.6(1) & 69.4(3)  & 69.7(3)                      \\
      he & 56.1 & 58.1(5)  & 60.6(2)  & 60.0(2)                      & 56.3 & 57.6(1) & 59.9(2)  & 58.8(5)                      \\
      \cmidrule{1-9}
      \multicolumn{9}{c}{\textit{Nearby}}                                                                                       \\
      \addlinespace
      bg & 69.3 & 71.9(2)  & 72.7(4)  & 72.5(4)                      & 75.0 & 75.9(2) & 76.7(2)  & 76.1(2)                      \\
      it & 81.5 & 80.1(2)  & 81.7(2)  & 81.5(1)                      & 74.7 & 62.8(7) & 84.8(3)  & 85.5(7)                      \\
      pt & 78.6 & 76.2(4)  & 78.1(3)  & 78.4(3)                      & 72.0 & 63.5(9) & 81.7(4)  & 83.2(10)                     \\
      fr & 80.0 & 81.3(2)  & 82.7(2)  & 82.8(1)                      & 65.7 & 54.9(5) & 71.7(7)  & 76.3(7)                      \\
      es & 71.8 & 72.0(6)  & 73.5(3)  & 73.7(2)                      & 67.8 & 58.1(2) & 73.9(7)  & 75.7(21)                     \\
      no & 68.4 & 74.1(2)  & 74.2(1)  & 74.4(2)                      & 62.2 & 61.4(2) & 64.7(5)  & 64.6(11)                     \\
      da & 67.5 & 70.4(4)  & 71.0(1)  & 70.9(3)                      & 72.9 & 72.0(1) & 76.0(3)  & 76.2(7)                      \\
      sv & 66.7 & 71.8(2)  & 72.1(1)  & 72.4(5)                      & 68.4 & 68.5(1) & 69.0(4)  & 70.7(6)                      \\
      nl & 64.8 & 66.9(2)  & 67.4(4)  & 68.8(4)                      & 72.9 & 70.3(3) & 74.3(4)  & 75.3(7)                      \\
      de & 57.2 & 64.0(9)  & 64.0(5)  & 63.9(5)                      & 52.8 & 59.3(3) & 58.9(5)  & 58.0(4)                      \\
      \bottomrule
    \end{tabular}
    \caption{Development set}
  \end{subtable}
  \begin{subtable}{1\textwidth}
    \centering
    \sisetup{table-number-alignment=right, separate-uncertainty, table-format=2.1(1)}
    \begin{tabular}{@{}cS[table-format=2.1]*{3}{S}S[table-format=2.1]*{3}{S}@{}}
      \toprule
      \multirow{2}{*}{Language} & \multicolumn{4}{c}{Parsing} & \multicolumn{4}{c}{Tagging} \\
      \cmidrule(lr){2-5} \cmidrule(l){6-9}
         & {MV} & {PPTX}   & {\gmn{}} & {\gmn{}, learned $\alpha_k$} & {MV} & {PPTX}   & {\gmn{}} & {\gmn{}, learned $\alpha_k$} \\
      \midrule
      \multicolumn{9}{c}{\textit{Distant}}                                                                                       \\
      \addlinespace
      fa & 43.7 & 42.5(10) & 49.4(4)  & 48.8(3)                      & 69.4 & 67.4(3)  & 71.1(7)  & 72.5(7)                      \\
      ar & 37.3 & 35.5(5)  & 37.3(5)  & 37.2(6)                      & 57.8 & 59.0(5)  & 63.0(14) & 63.3(16)                     \\
      id & 59.0 & 57.4(6)  & 61.6(2)  & 61.4(2)                      & 77.9 & 79.9(4)  & 81.0(2)  & 80.8(8)                      \\
      ko & 14.7 & 14.8(5)  & 13.8(4)  & 14.7(1)                      & 45.0 & 45.7(2)  & 45.9(11) & 44.9(15)                     \\
      tr & 20.1 & 19.3(2)  & 19.7(3)  & 20.8(2)                      & 62.8 & 63.0(2)  & 64.7(5)  & 64.9(1)                      \\
      hi & 21.1 & 23.0(5)  & 21.7(2)  & 26.4(3)                      & 59.7 & 65.1 (2) & 62.8(5)  & 65.1(10)                     \\
      hr & 57.4 & 62.2(2)  & 64.3(4)  & 64.2(3)                      & 66.7 & 58.4(1)  & 68.5(2)  & 69.0(4)                      \\
      he & 56.2 & 57.5(7)  & 60.1(2)  & 59.7(3)                      & 55.5 & 57.2(1)  & 59.1(4)  & 58.0(4)                      \\
      \cmidrule{1-9}
      \multicolumn{9}{c}{\textit{Nearby}}                                                                                        \\
      \addlinespace
      bg & 69.3 & 72.4(2)  & 73.1(3)  & 72.9(3)                      & 75.6 & 76.2(1)  & 77.1(3)  & 76.6(2)                      \\
      it & 81.7 & 80.2(1)  & 81.4(2)  & 81.4(3)                      & 74.5 & 62.1(8)  & 85.0(4)  & 86.0(8)                      \\
      pt & 76.6 & 74.5(4)  & 76.3(3)  & 76.5(3)                      & 72.5 & 63.4(8)  & 81.8(5)  & 83.1(9)                      \\
      fr & 76.5 & 78.2(3)  & 79.3(1)  & 79.1(1)                      & 65.4 & 56.0(4)  & 72.2(6)  & 75.7(5)                      \\
      es & 71.3 & 71.5(6)  & 73.1(4)  & 73.2(3)                      & 67.1 & 57.8(2)  & 73.5(6)  & 75.1(20)                     \\
      no & 69.1 & 74.1(2)  & 74.2(2)  & 74.5(2)                      & 63.2 & 62.3(3)  & 65.9(6)  & 65.7(11)                     \\
      da & 67.3 & 70.7(3)  & 71.3(1)  & 71.2(4)                      & 73.9 & 72.8(1)  & 77.1(4)  & 77.3(6)                      \\
      sv & 70.1 & 74.5(3)  & 74.7(2)  & 75.0(3)                      & 69.8 & 69.6(1)  & 70.0(3)  & 72.0(5)                      \\
      nl & 65.8 & 67.8(4)  & 68.5(3)  & 69.6(4)                      & 70.8 & 68.7(3)  & 73.0(3)  & 74.5(10)                     \\
      de & 55.3 & 61.6(8)  & 61.9(6)  & 61.7(5)                      & 50.2 & 56.4(3)  & 55.8(6)  & 54.9(4)                      \\
      \bottomrule
    \end{tabular}
    \caption{Test set}
  \end{subtable}
  \caption{Full performance results. Numbers are averages ($\pm$ std) over 5
    runs with different random seeds. For parsing, the numbers correspond to
    labelled attachment score (LAS) whereas for tagging, they correspond to
    accuracy. Both metrics are better if higher.}\label{tbl:full-results}
\end{table*}

\end{document}